\documentclass[letterpaper]{article} % DO NOT CHANGE THIS
\usepackage{aaai24}  % DO NOT CHANGE THIS
\usepackage{times}  % DO NOT CHANGE THIS
\usepackage{helvet}  % DO NOT CHANGE THIS
\usepackage{courier}  % DO NOT CHANGE THIS
\usepackage[hyphens]{url}  % DO NOT CHANGE THIS
\usepackage{graphicx} % DO NOT CHANGE THIS
\usepackage{natbib}  % DO NOT CHANGE THIS AND DO NOT ADD ANY OPTIONS TO IT
\usepackage{caption} % DO NOT CHANGE THIS AND DO NOT ADD ANY OPTIONS TO IT
\usepackage{algorithm}
\usepackage{algorithmic}
\usepackage{multicol,multirow}
\usepackage{booktabs}
\usepackage{amsfonts}       % blackboard math symbols
\usepackage{amsmath}

\usepackage{newfloat}
\usepackage{listings}
\DeclareCaptionStyle{ruled}{labelfont=normalfont,labelsep=colon,strut=off} % DO NOT CHANGE THIS
\floatstyle{ruled}
\newfloat{listing}{tb}{lst}{}
\floatname{listing}{Listing}

\title{Controllable Mind Visual Diffusion Model}
\author{
    Bohan Zeng{$^{1}$}\thanks{These authors contributed equally.}, Shanglin Li{$^1$}\footnotemark[1], Xuhui Liu{$^1$}, Sicheng Gao{$^{1}$} \\ \textbf{Xiaolong Jiang{$^{3}$}, Xu Tang{$^{3}$}, Yao Hu{$^{3}$}, Jianzhuang Liu{$^2$}, Baochang Zhang{$^{1,4,5}$}\thanks{Corresponding Author: bczhang@buaa.edu.cn.} }\\
}
\affiliations{
    \textsuperscript{\rm 1}Institute of Artificial Intelligence, Hangzhou Research Institute, Beihang University, China \\ \textsuperscript{\rm 2}Shenzhen Institute of Advanced Technology, Shenzhen, China \ \textsuperscript{\rm 3}Xiaohongshu Inc \\ \textsuperscript{\rm 4} Zhongguancun Laboratory, Beijing, China \ \textsuperscript{\rm 5}Nanchang Institute of Technology, Nanchang, China\\
}

\begin{document}

\maketitle

\begin{abstract}

Brain signal visualization has emerged as an active research area, serving as a critical interface between the human visual system and computer vision models. Diffusion-based methods have recently shown promise in analyzing functional magnetic resonance imaging (fMRI) data, including the reconstruction of high-quality images consistent with original visual stimuli. Nonetheless, it remains a critical challenge to effectively harness the semantic and silhouette information extracted from brain signals. In this paper, we propose a novel approach, termed as Controllable Mind Visual Diffusion Model (CMVDM). Specifically, CMVDM first extracts semantic and silhouette information from fMRI data using attribute alignment and assistant networks. Then, a control model is introduced in conjunction with a residual block to fully exploit the extracted information for image synthesis, generating high-quality images that closely resemble the original visual stimuli in both semantic content and silhouette characteristics. Through extensive experimentation, we demonstrate that CMVDM outperforms existing state-of-the-art methods both qualitatively and quantitatively. \textbf{Our code is available\footnote{https://github.com/zengbohan0217/CMVDM}.}
\end{abstract}

\section{Introduction}

\begin{figure}[t]
	\centering
	\hspace{-3.5mm}
	\includegraphics[width=0.46\textwidth]{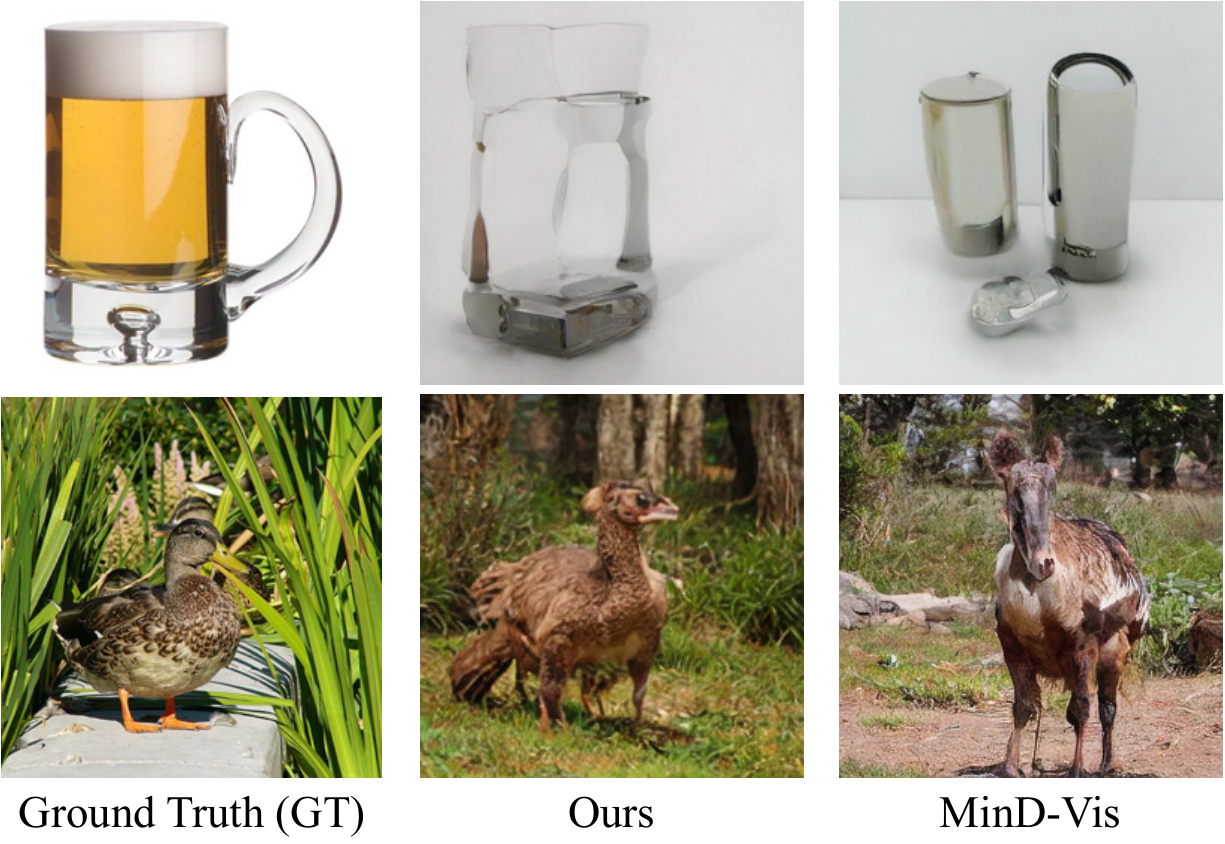} \\
	\caption{Illustration of synthesis results. A recent method MinD-Vis \cite{chen2022seeing} can generate photo-realistic results, but they cannot well match the visual stimuli in terms of semantics and silhouette. Our method can generate better results more consistent with the GT visual stimuli.} 
	\label{first image}
\end{figure}

Understanding the cognitive processes that occur in the human brain when observing visual stimuli (e.g., natural images) has long been a primary focus for neuroscientists. Both objective visual stimuli and subjective cognitive activities can elicit the transmission of intricate neural signals in the visual cortex of the brain, thus laying the foundation for higher-order cognitive and decision-making processes. With the advancement of techniques such as functional magnetic resonance imaging (fMRI), it has become possible to capture real-time brain activity signals with greater accuracy and finer granularity, thereby accelerating the progress of neuroscientific research. Deciphering and reconstructing from these intricate signals remain a great challenge to both cognitive neuroscience and downstream applications like Brain-Computer Interfaces (BCI) \cite{nicolas2012brain,milekovic2018stable}.

Early attempts \cite{van2010efficient, damarla2013decoding, horikawa2017generic, akamatsu2020brain} at analyzing brain activity on visual tasks mainly focus on matching human subjects' brain activity with observed natural images, or reconstructing visual patterns of simple geometric shapes \cite{miyawaki2008visual, schoenmakers2013linear, van2010neural}. These explorations demonstrate the feasibility of deriving semantic information for perceived images from brain signals, yet they have poor generalization to unseen semantic categories or complicated reconstruction tasks.

Recent studies \cite{beliy2019voxels, gaziv2022self, ozcelik2022reconstruction, chen2022seeing, takagi2022high} have made significant progress in reconstructing visual stimuli from brain signals. \cite{beliy2019voxels, gaziv2022self} can generate images that are similar in shape to the original visual stimuli, but the images suffer from severe distortion and blur issues. \cite{ozcelik2022reconstruction, chen2022seeing, takagi2022high} have employed commonly used generative models, such as Generative Adversarial Networks (GAN) or diffusion models, to generate high-quality RGB images that maintain semantic consistency with the original visual stimuli conditioned on corresponding fMRI signals. However, such methods struggle with positional inconsistency, as shown in Fig. \ref{first image}. 
In general, existing methods have not effectively utilized the semantic and spatial features inherent in fMRI signals.

In this paper, we present a Controllable Mind Visual Diffusion Model (CMVDM) that enables the mind diffusion model with a control network to leverage the extracted faithful semantic and silhouette information for high-fidelity human vision reconstruction.
Specifically, we first finetune a pretrained latent diffusion model (LDM) with a semantic alignment loss and pretrain a silhouette extractor to estimate accurate semantic and silhouette information of the fMRI data. Taking inspiration from ControlNet, we then introduce a control network, which takes the silhouette information as a condition, into the pretrained LDM to guide the diffusion process to generate desired images that match the original visual stimuli in terms of both semantic and silhouette information.
Fig. \ref{first image} shows two examples where CMVDM outperforms the previous state-of-the-art approach, MinD-Vis. 

In summary, the main contributions of this paper are as follows:

\begin{itemize}

    \item 
    We propose a novel Controllable Mind Visual Diffusion Model (CMVDM) that leverages both semantic and spatial visual patterns in brain activity to reconstruct photo-realistic images. A control network is utilized to enable effective manipulation over the positions of generated objects or scenes in the reconstructed images, providing a much better structural similarity to the original visual stimuli.

    \item 
    We design two extractors to extract semantic and silhouette attributes to provide accurate information for generating images that closely resemble the visual stimuli. Besides, we build a residual module to provide information beyond semantics and silhouette. 

    \item
    We conduct comprehensive experiments on two datasets to evaluate the performance of our method. It achieves state-of-the-art qualitative and quantitative results compared to existing methods, demonstrating the efficacy of CMVDM for decoding high-quality and controllable images from fMRI signals.

\end{itemize}

\begin{figure*}[t]
	\centering
	\hspace{-3.5mm}
	\includegraphics[width=0.98\textwidth]{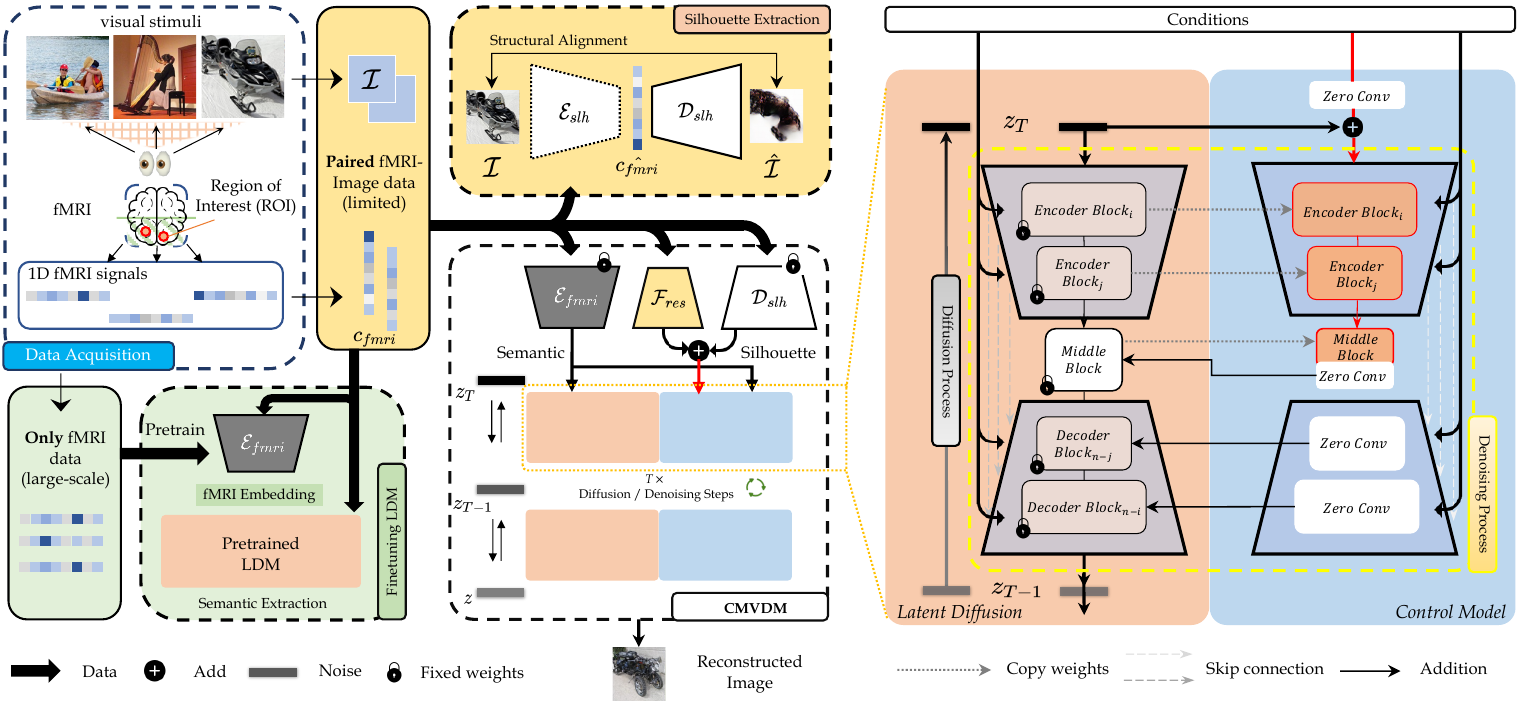} \\
     \caption{Overview of our proposed method. Initially, we train $\mathcal{E}_{fmri}$ and $\mathcal{D}_{slh}$ in the ``Finetuning LDM'' and ``Silhouette Extraction'' parts, respectively. Subsequently, we utilize $\mathcal{E}_{fmri}$, $\mathcal{D}_{slh}$, and $\mathcal{F}_{res}$ to extract semantic, silhouette, and supplement information from fMRI signals as conditions. Finally, we integrate the control network with the LDM to generate high-fidelity and controllable results tailored to the aforementioned conditions.} 

	\label{framework}
\end{figure*}

\section{Related Work}
\label{related_work}

\paragraph{Diffusion Probabilistic Models.}

Diffusion models (DMs) were initially introduced by \cite{sohl2015deep} as a novel generative model that gradually denoises images corrupted by Gaussian noise to produce samples. Recent advances in DMs have demonstrated their superior performance in image synthesis, with notable models including \cite{ho2020denoising, song2020denoising, dhariwal2021diffusion, vahdat2021score, rombach2022high, peebles2022scalable}. DDGAN \cite{xiao2022tackling} is a model that reduces the number of sampling steps by directly predicting the ground truth in each timestep. DMs have also achieved state-of-the-art performance in other synthesis tasks, such as text-to-image generation with GLIDE \cite{nichol2021glide}, speech synthesis with \cite{kong2020diffwave, liu2021diffsinger}, and super-resolution with \cite{li2022srdiff, saharia2022image, gao2023implicit}. In addition, DMs have been applied to text-to-3D synthesis in \cite{poole2022dreamfusion, lin2022magic3d}, and other 3D object syntheses in \cite{anciukevivcius2022renderdiffusion, li2022diffusion, luo2021diffusion}. Furthermore, DMs have found applications in video synthesis \cite{ho2022video, ho2022imagen}, semantic segmentation \cite{baranchuk2021label}, text-to-motion generation \cite{tevet2022human}, face animation \cite{zeng2023face}, and object detection \cite{chen2022diffusiondet}. \cite{kulikov2022sinddm, wang2022sindiffusion} are models that generate diverse results by learning the internal patch distribution from a single image. ControlNet employs a control network on a pretrained text-conditioned LDM for controllable image synthesis. Overall, DMs have shown promising results and have been widely adopted in various synthesis tasks.

\paragraph{Neural Decoding of Visual Stimuli.}
Neural decoding of visual stimuli has been a topic of growing interest in recent years. Numerous studies have explored the possibility of using machine learning algorithms to decode visual information from patterns of neural activity in the human brain. For instance, \cite{naselaris2009bayesian} demonstrates that it is possible to reconstruct natural images from fMRI data using a linear decoder. Similarly, \cite{kay2008identifying} shows that the orientation of gratings from patterns of activity in the early visual cortex can be decoded using a support vector machine. More recent studies have built on these findings by exploring more complex visual stimuli, such as natural scenes \cite{nishimoto2011reconstructing} and faces \cite{kriegeskorte2007individual}, and by developing more sophisticated machine learning algorithms, such as deep neural networks \cite{yamins2014performance}. To enable decoding of novel scenarios, some works use an identification-based approach \cite{horikawa2017generic, akamatsu2020brain, kay2008identifying}, where they model the relationship between brain activity and visual semantic knowledge such as image features extracted by a CNN \cite{horikawa2017generic,akamatsu2020brain}. These studies provide valuable insights into the interpretation of human brain signals in the visual cortex, which can help the development of more effective decoding algorithms for a wide range of neuroimaging applications, such as Brain-Computer Interfaces. However, these methods require a large amount of paired stimuli-responses data that is hard to obtain. Therefore, decoding novel image categories accurately remains a challenge.

\paragraph{fMRI-to-Image Reconstruction}
With the remarkable advancements in generative models, recent studies have focused on the reconstruction of images from human brain activity. These studies employ various approaches, such as building an encoder-decoder structure to align image features with corresponding fMRI data, as demonstrated by \cite{beliy2019voxels} and \cite{gaziv2022self}. To further enhance the quality of image reconstruction, researchers have turned to more sophisticated techniques, including generative adversarial networks (GAN) \cite{ozcelik2022reconstruction} and diffusion models \cite{takagi2022high,chen2022seeing}. These methods have shown promise in achieving more plausible image reconstruction. Nonetheless, the approaches described above have limitations in terms of image reconstruction quality and localization accuracy, resulting in unreliable reconstruction outcomes and inadequate utilization of the deep semantic and shallow positional information inherent in fMRI signals.

\section{Method}
\label{methods}

In this section, we describe the CMVDM model, which combines attribute extractors and a control model to produce precise and controllable outcomes from fMRI signals. Fig. \ref{framework} illustrates the architecture of CMVDM.

\subsection{Problem Statement and Overview of CMVDM}

Let the paired \{fMRI, image\} dataset $\Omega = \{(c_{fmri,i}, \mathcal{I}_{i})\}_{i=1}^n$, where $c_{fmri,i} \in \mathbb{R}^{1 \times N}$ and $\mathcal{I}_{i} \in \mathbb{R}^{H \times W \times 3}$. The fMRI data is extracted as a 1D signal from the region of interest (ROI) on the visual cortex averaged across the time during which the visual stimuli are presented.  $N$ denotes the number of voxels of the extracted signal.
We adopt the pretrained image encoder of the LDM \cite{rombach2022high} to encode the observed image $\mathcal{I}$ into the latent code $z$. Our CMVDM aims to learn an estimation of the data distribution $p(z | c_{fmri})$ through a Markov chain with $T$ timesteps. Following \cite{ho2020denoising, song2020denoising, rombach2022high}, we define the fixed forward Markov diffusion process $q$ as:
\begin{equation}
\small
    \begin{aligned}
    q\left(z_{1: T} \mid z_0\right) &= \prod_{t=1}^T q\left(z_t \mid z_{t-1}\right), \\
    q\left(z_t \mid z_{t-1}\right) &= \mathcal{N}\left(z_t \mid \sqrt{1-\beta_t} z_{t-1},\beta_t \mathbf{I}\right), \\
    \end{aligned}
\end{equation}

where $z_0$ denotes the latent code of an image. This Markov diffusion process propagates by adding Gaussian noise, with variances $\beta_t \in (0,1)$ in $T$ iterations. Given $z_{0}$, the distribution of $z_{t}$ can be represented by: 
\begin{equation}
\small
\begin{aligned}
q\left(z_t \mid z_{0}\right) &=\mathcal{N}\left(z_t \mid \sqrt{\gamma_t} z_{0}, (1-\gamma_t) \mathbf{I}\right),
\end{aligned}
\end{equation}

where $\gamma_t = \prod_{i=1}^t \left(1 - \beta_i\right)$. In the inference process, CMVDM learns the conditional distributions $p_{\theta}(z_{t-1} | z_t, c_{fmri})$ and conducts a reverse Markov process from Gaussian noise $z_T \sim \mathcal{N}(\mathbf{0}, \mathbf{I})$ to a target latent code $z_0$ as:
\begin{equation}
\small
\begin{aligned}
p_\theta\left(z_{0: T} \mid c_{fmri}\right) &=p\left(z_T\right) \prod_{t=1}^T p_\theta\left(z_{t-1} \mid z_t, c_{fmri}\right), \\
p\left(z_T\right) &=\mathcal{N}\left(z_T \mid \mathbf{0}, \mathbf{I}\right), \\
p_\theta\left(z_{t-1} \mid z_t, c_{fmri} \right) &=\mathcal{N}\left(z_{t-1} \mid \mu_\theta\left(c_{fmri}, z_t, t\right), \sigma_t^2 \mathbf{I}\right),
\end{aligned}
\end{equation}

where $\sigma_t = \frac{1 - \gamma_{t-1}}{1 - \gamma_t} \beta_t$. The pretrained image decoder of the LDM \cite{rombach2022high} turns the final latent code to an image.

Furthermore, we extract the attributes and control the generated results. Firstly, we extract the semantic and silhouette information by utilizing the fMRI encoder $\mathcal{E}_{fmri}$ and the silhouette estimating network $\mathcal{D}_{slh}$, respectively. This step enables us to accurately decouple the fMRI information $c_{fmri}$. Subsequently, we utilize the control model $\mathcal{F}_{ctrl}$ to generate high-quality images that match the visual stimuli in terms of both semantic and silhouette information. $\mathcal{F}_{ctrl}$ is able to leverage the extracted information to produce better results. Besides, the residual module $\mathcal{F}_{res}$ is designed to provide information beyond semantics and silhouette.

\subsection{Finetuning of the Pretrained LDM}
\label{stage_one}

Before extracting the silhouette information and controlling the generated results, we need to finetune the pretrained LDM  \cite{rombach2022high} to enable it to generate consistent images and extract the semantic information based on the input fMRI signals.
Following MinD-Vis, we employ the fMRI encoder $\mathcal{E}_{fmri}$ pretrained on the HCP dataset \cite{van2013wu} to encode the brain activity signals to the fMRI embeddings. Besides, we use the pretrained LDM to generate output images. By optimizing the fMRI encoder $\mathcal{E}_{fmri}$ and the cross-attention layers in the LDM, while freezing the other blocks during the finetuning process, we can obtain reliable consistent generated results. The finetuning loss is defined as follows:
\begin{equation}
    \begin{aligned}
        \mathcal{L}_{f} &= \mathbb{E}_{z_0, t, c_{fmri}, \epsilon \sim \mathcal{N}(0,1)} [ || \epsilon - \epsilon_{\theta}(z_t, t, \mathcal{E}_{fmri}(c_{fmri})) ||^2_2 ],
    \end{aligned}
\end{equation}

where $\epsilon_{\theta}$ is the denoising network of the LDM. In this way, 
the LDM can ensure the consistency of the generated results.
Let $c_{ctx} = \mathcal{E}_{fmri}(c_{fmri})$ be the semantic information extracted from the fMRI signals.
Due to the lack of direct semantic supervision, $\mathcal{E}_{fmri}$ may be insufficient for providing enough semantic information. Therefore, we design a noval alignment loss $\mathcal{L}_{align}$ to further enhance the semantic information $c_{ctx}$:
\begin{equation}
    \begin{aligned}
        \mathcal{L}_{align} &= e^{-{\rm cosine}(f_{img}, {\rm MLP}(c_{ctx}))},
    \end{aligned}
\end{equation}

where ${\rm cosine}(\cdot, \cdot)$ denotes the cosine similarity, $f_{img}$ is the image feature extracted by the CLIP image encoder \cite{radford2021learning}, and ${\rm MLP}$ represents a trainable multi-layer perceptron.
After this training stage, the LDM can make the generated images consistent with the fMRI signals. Nonetheless, due to the absence of explicit positional condition guidance, it is still a challenge for the LDM to generate silhouette-matched results. In the next two sections, we will describe how to extract silhouette information from the fMRI signals and control the final results.

\subsection{Silhouette Extraction}
\label{Silhouette Extraction}

In this section, we aim to extract silhouette information from fMRI signals. 
\cite{gaziv2022self} uses a combination of self-supervised and supervised learning to reconstruct images similar to visual stimuli.

Despite the low fidelity of the image generation quality, their generated results demonstrate a notable ability to accurately replicate the silhouette of the visual stimuli (see Fig. \ref{comparison_GOD}). Based on this, we devise a silhouette estimation network that is capable of providing rough positional guidance for CMVDM.

Our silhouette estimation network consists of two components: an encoder $\mathcal{E}_{slh}$ and a decoder $\mathcal{D}_{slh}$. The encoder $\mathcal{E}_{slh}$ projects the input images to the fMRI signal space, while the decoder $\mathcal{D}_{slh}$ performs the inverse transformation.

Let $c_{fmri,i}$ be the ground truth (GT) fMRI signal, $\mathcal{I}_{i}$ be the corresponding GT image, and $\hat{c_{fmri,i}} = \mathcal{E}_{slh}(\mathcal{I}_{i})$ be the estimated fMRI signal. We define the encoder training loss $\mathcal{L}_{e}$ by a combination of the Mean Square Error (MSE) loss and cosine similarity:
\begin{equation}
\small
    \begin{aligned}
       % \mathcal{L}_{cos, i} &= 1 - {\rm cosine}(c_{fmri,i}, \hat{c_{fmri,i}}), \ i=1,2,..., |\Omega|, \\
       \mathcal{L}_{e} &= \frac{1}{|\Omega|}\sum_{i=1}^{|\Omega|} [\alpha_{1} \cdot  \left \| c_{fmri,i}- \hat{c_{fmri,i}} \right \|^2 \\
       & + \alpha_{2} \cdot (1 - {\rm cosine}(c_{fmri,i}, \hat{c_{fmri,i}})) ],
    \end{aligned}
\end{equation}

where $\alpha_{i \in \{1,2\}}$ are the hyperparameters set empirically to $\alpha_{1}=1$ and $\alpha_{2}=0.3$.

After completing the training of $\mathcal{E}_{slh}$, we fix its parameters and train the reverse process for the decoder $\mathcal{D}_{slh}$.
Due to the limited availability of paired \{fMRI, image\} data, mapping fMRI signals to images is challenging. 
Inspired by \cite{gaziv2022self}, we utilize semi-supervised training to extract intricate silhouette information.
The self-supervised process can be simply represented as: $\hat{\phi_{i}} = \mathcal{D}_{slh}(\mathcal{E}_{slh}(\phi_{i}))$, where $\phi_{i} \in \Phi$ denotes the image from ImageNet (without corresponding fMRI data) \cite{deng2009imagenet}, and $\hat{\phi_{i}}$ denotes the reconstructed image.
By minimizing the disparity between $\phi_{i}$ and $\hat{\phi_{i}}$, the self-supervised process helps $\mathcal{E}_{slh}$ and $\mathcal{D}_{slh}$ to learn more generalized image representation.
We employ the Structural Similarity (SSIM) loss besides the Mean Absolute Error (MAE) loss to penalize the spatial distances between the reconstructed images and the GT images. The two losses are:
% \begin{gather*}
\begin{equation}
\small
    \begin{aligned}
    \mathcal{L}_{mae} =&  
    \underbrace{\frac{1}{|\Omega|}\sum_{i=1}^{|\Omega|}|\hat{\mathcal{I}_{i}}-\mathcal{I}_{i}|}_{supervised} 
    + \underbrace{\frac{1}{|\Phi|}\sum_{i=1}^{|\Phi|}|\hat{\phi_{i}}-\phi_{i}|}_{self-supervised}, \\
    % & + \beta_{3} \cdot \sum_{i=1}^{|\Psi|}|\psi_{i}|,
    \end{aligned}
\end{equation}
\begin{equation}
\small
    \begin{aligned}
    \mathcal{L}_{ssim} = 1 - \frac{(2\mu_{\mathcal{I}}\mu_{\hat{\mathcal{I}}} + C_1)(2\sigma_{\mathcal{I}\hat{\mathcal{I}}} + C_2)}{(\mu_{\mathcal{I}}^2 + \mu_{\hat{\mathcal{I}}}^2 + C_1)(\sigma_{\mathcal{I}}^2 + \sigma_{\hat{\mathcal{I}}}^2 + C_2)},
    \end{aligned}
\end{equation}

where $\mu_{\hat{\mathcal{I}}}$, $\mu_{\mathcal{I}}$, $\sigma_{\hat{\mathcal{I}}}$, and $\sigma_{\mathcal{I}} $ represent the mean and std values of the reconstructed images $\hat{\mathcal{I}}$ and GT images $\mathcal{I}$, $C_1$ and $C_2$ are constants to stabilize the calculation.

The decoder loss $\mathcal{L}_{d}$ is defined as the combination of the two losses:
\begin{equation}
    \mathcal{L}_{d} = \mathcal{L}_{mae} + \mathcal{L}_{ssim}. 
\end{equation}

After training, $\mathcal{D}_{slh}$ is able to generate images $\hat{\mathcal{I}}$ from $c_{fmri}$ that provide positional guidance for CMVDM. To avoid confusion, we'll refer to $\hat{\mathcal{I}}$ as $c_{slh}$ in the following section.

\begin{table*}[]
\centering
\caption{Quantitative comparison with four state-of-the-art (SOTA) methods. Bold results denote the best results and underlined results denote the second-best results.}
\label{Comparison Table}
\begin{tabular}{lcccccc}
\hline
\multirow{2}{*}{Method} & \multicolumn{3}{c}{GOD}                                                                                      & \multicolumn{3}{c}{BOLD5000}                                                                                \\ \cmidrule(r){2-4} \cmidrule(r){5-7} 
                        & \multicolumn{1}{c}{Acc (\%)} & \multicolumn{1}{c}{PCC} & \multicolumn{1}{c}{SSIM} & \multicolumn{1}{c}{Acc(\%)} & \multicolumn{1}{c}{PCC} & \multicolumn{1}{c}{SSIM} \\ \hline
Beliy (2019)                   & 4.288  & 0.48285  & 0.51795   & /   & /  & /    \\
Gaziv (2022)                   & 9.128  & \underline{0.68326}  & \textbf{0.64857}   & /   & /  & /    \\
IC-GAN (2022)                  & \underline{29.386}  & 0.44857  & 0.54489   & /  & /  & /    \\
MinD-Vis (2023)                & 26.644  & 0.53159  & 0.52669   & \underline{25.918}  & \underline{0.54486}  & \underline{0.52379}   \\ \hline
CMVDM (Ours)                    & \textbf{30.112}  & \textbf{0.76751}  & \underline{0.63167}   & \textbf{27.791}  & \textbf{0.55691}  & \textbf{0.53459}   \\ \hline
\end{tabular}
\end{table*}

\subsection{Training of Control Model}

After obtaining the enhanced semantic information $c_{ctx}=\mathcal{E}_{fmri}(c_{fmri})$ and the reliable silhouette information $c_{slh}=\mathcal{D}_{slh}(c_{fmri})$ from $c_{fmri}$, we use them to control the generated results as shown in Fig. \ref{framework}. 
Inspired by ControlNet, we design a control model to control the overall composition of the generated images. Specifically, we freeze all the parameters in the denoising network $\epsilon_{\theta}$ and clone the U-Net encoder of $\epsilon_{\theta}$ into the trainable $\mathcal{F}_{ctrl} (\cdot; \Theta_c)$ with a set of parameters $\Theta_c$ (the red blocks of control model in Fig. \ref{framework}). The inputs of $\mathcal{F}_{ctrl}$ include $z_t$, $c_{ctx}$, and the silhouette feature $c_{slh}$. The combined condition code $x^\prime_{c, t}$ can be formulated as:
\begin{equation}
    \begin{aligned}
        x^\prime_{c, t} &= \mathcal{Z}( \mathcal{F}_{ctrl}(z_t + \mathcal{Z}(c_{slh}), c_{ctx}; \Theta_c ) ),
    \end{aligned}
\end{equation}

where $\mathcal{Z}(\cdot)$ denotes the zero convolution operation \cite{zhang2023adding}. Furthermore, in order to compensate for the fMRI data loss during attribute extraction, we utilize a trainable residual block denoted as $\mathcal{F}_{res}$. This block is trained in conjunction with $\mathcal{F}_{ctrl}$. The final combined condition code $x_{c, t}$ is represented as:
\begin{equation}
    \begin{aligned}
        x_{c, t} =& \mathcal{Z}( \mathcal{F}_{ctrl}(z_t + \\
        & \mathcal{Z}(c_{slh}+\mathcal{Z}(\mathcal{F}_{res}(c_{fmri}))), c_{ctx}; \Theta_c ) ).
    \end{aligned}
\end{equation}

Then the output features $x_{c, t}$ of the control model are added to the U-Net decoder features of the frozen $\epsilon_{\theta}$, as shown in Fig. \ref{framework}.

Finally, we use the following loss $\mathcal{L}_{ctrl}$ to supervise the training of the control model and $\mathcal{F}_{res}$ in our CMVDM:
\begin{equation}
    \begin{aligned}
        % \epsilon_{pred} &= \mathcal{F}_{ctrl} (z_t, t, \mathcal{E}_{fmri}(c_{fmri}), \mathcal{D}_{slh}(c_{fmri}))      \\
        \mathcal{L}_{ctrl} &=  \\
        & \mathbb{E}_{z_0, t, c_{fmri}, \epsilon \sim \mathcal{N}(0,1)} [ || \epsilon - \epsilon_{\theta}(z_t, t, c_{ctx}, x_{c,t}) ||^2_2 ]. \\
    \end{aligned}
\end{equation}

Note that with their losses, the control model training, the pretrained LDM finetuning, and the $\mathcal{D}_{slh}$ training are independent. In our framework, we separately pretrained $\mathcal{E}_{fmri}$ and $\mathcal{D}_{slh}$ and froze their weights to jointly train $\mathcal{F}_{res}$ and $\mathcal{F}_{ctrl}$ (as depicted in Fig \ref{framework}).

\begin{figure*}[t]
	\centering
	\hspace{-3.5mm}
	\includegraphics[width=0.86\textwidth]{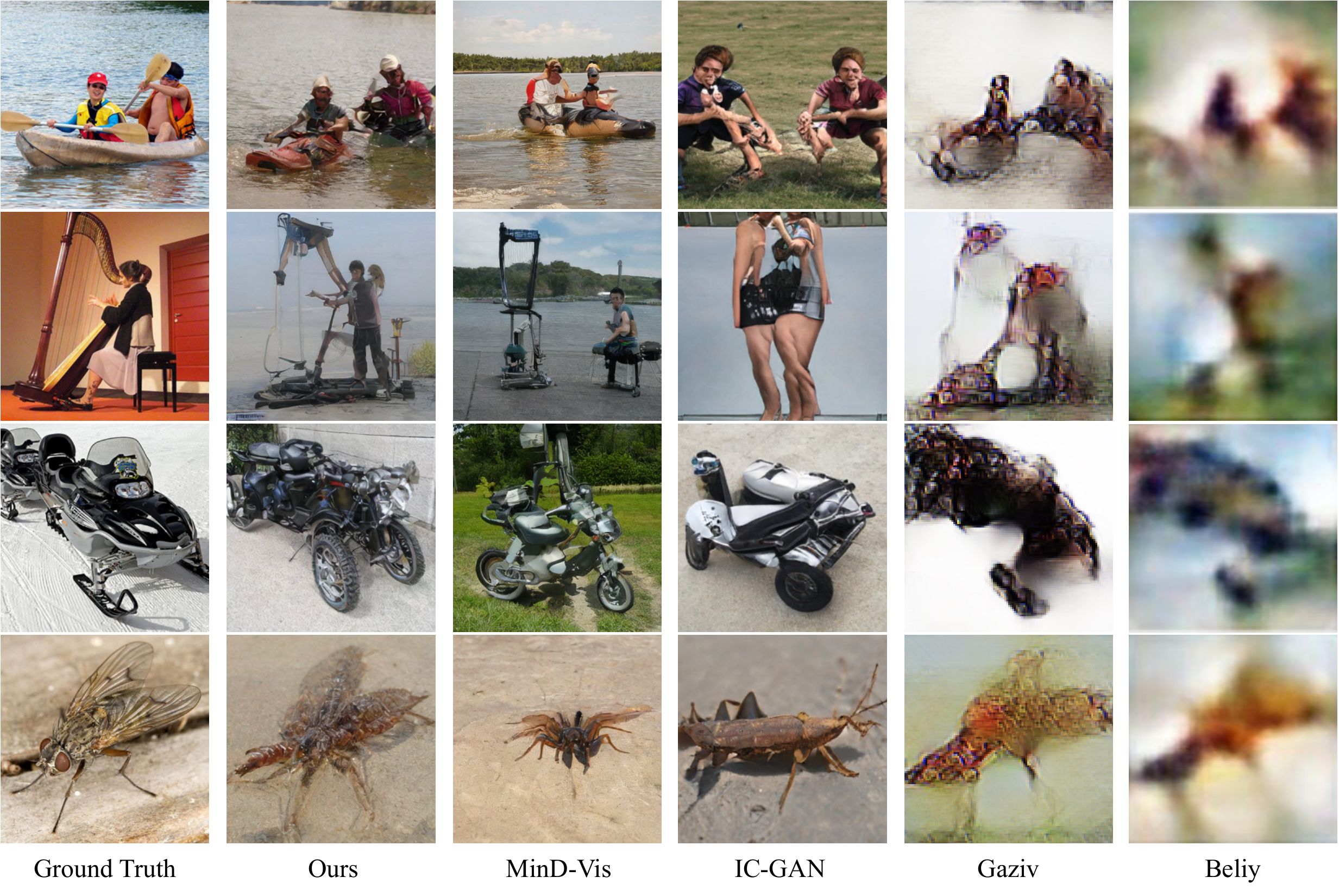} \\
	\caption{Comparison with four SOTA methods on the GOD dataset.} 
	\label{comparison_GOD}
	% \vspace{-5mm}
\end{figure*}

\begin{figure}[t]
	\centering
	\hspace{-3.5mm}
	\includegraphics[width=0.43\textwidth]{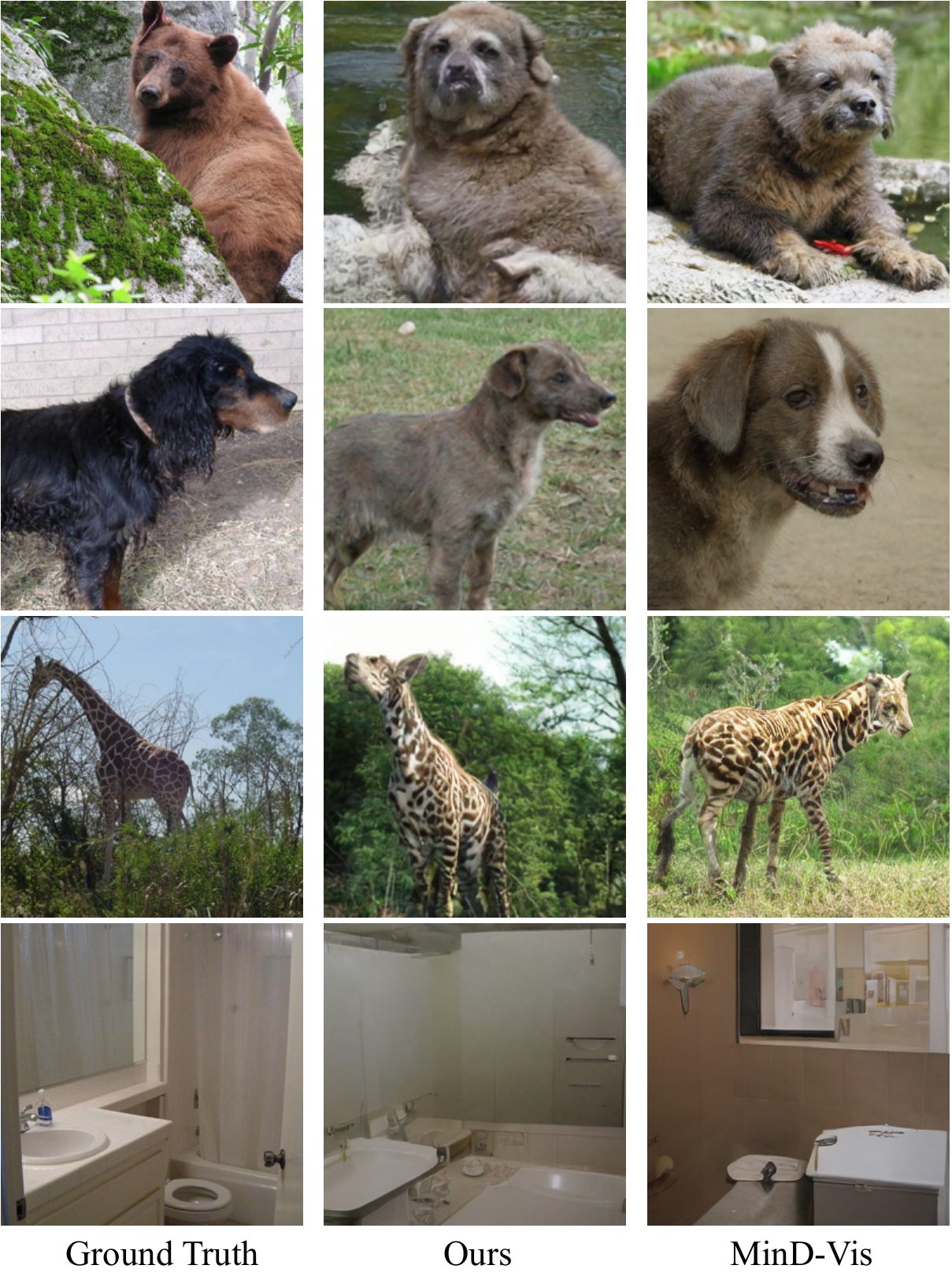} \\
	\caption{Comparison with MinD-Vis on the BOLD5000 dataset.} 
	\label{comparison_BOLD5000}
\end{figure}

\section{Experiments}
\label{experiments}

\subsection{Datasets and Implementation}

\paragraph{Datasets.}
In this study, we employ two public datasets with paired fMRI signals and images:
Generic Object Decoding (GOD) dataset \cite{horikawa2017generic}, and Brain, Object, Landscape Dataset (BOLD5000) \cite{chang2019bold5000}. The GOD dataset is a well-known and extensively researched collection of fMRI-based brain signal decoding data. It comprises 1250 distinct images belonging to 200 different categories, with 50 images designated for testing.
The BOLD5000 dataset is a rich resource for studying the neural representation of visual stimuli, as it contains diverse images from natural and artificial domains. The images are drawn from three existing datasets: SUN \cite{xiao2010sun}, COCO \cite{lin2014microsoft}, and ImageNet \cite{deng2009imagenet}, which contain images of various categories of objects and animals. BOLD5000 was acquired from four subjects who underwent fMRI scanning while viewing 5,254 images in 15 sessions. The fMRI data were preprocessed and aligned to a common anatomical space, resulting in 4803 fMRI-image pairs for training and 113 for testing. The dataset provides a unique opportunity to investigate how the human brain encodes visual information across different levels of abstraction and complexity.
Additionally, we use the large-scale fMRI data from Human Connectome Project (HCP) \cite{van2013wu} in an unsupervised manner to pretrain the fMRI encoder $\mathcal{E}_{fmri}$ in our method, which aims to fully extract the features of fMRI signals.

\paragraph{Training Details.}

We adopt 1 A100-SXM4-40GB GPU for the training of $\mathcal{E}_{fmri}$ and the control model, and 1 V100-SXM2-32GB GPU for $\mathcal{D}_{slh}$ training. 
Both $\mathcal{E}_{fmri}$ and the control model are trained by the AdamW \cite{loshchilov2017decoupled} with $\beta = (0.9, 0.999)$ and $eps = 1e-8$ for 500 epochs.
$\mathcal{D}_{slh}$ is optimized using Adam \cite{kingma2014adam} with a learning rate of $5e-3$ and $\beta=(0.5, 0.99)$ for 150 epochs.

\subsection{Evaluation Metrics}

\paragraph{N-way Classification Accuracy (Acc).}
Following \cite{gaziv2022self, chen2022seeing}, we employ the $n$-way top-1 classification task to evaluate the semantic correctness of the generated results, where multiple trials for top-1 classification accuracies are calculated in $n-1$ randomly selected classes with the correct class. Specifically, we follow MinD-Vis and use a pretrained ImageNet-1K classifier \cite{dosovitskiy2020image} to estimate the accuracy. Firstly, we input the generated results and the ground-truth images into the classifier, and then check whether the top-$1$ classification matches the correct class. \textit{More details about this metric can be found in our supplementary material}.

\paragraph{Pearson Correlation Coefficient (PCC).}
The Pearson correlation coefficient (PCC) measures the degree of linear association between two variables. PCC is used to measure the correlation between the pixel values of the generated results and those of the ground truth, with +1 indicating a perfect positive linear relationship and -1 indicating a perfect negative linear relationship. The larger the PCC value, the stronger the relevance between visual stimuli and generated images.

\paragraph{Structure Similarity Index Measure (SSIM).}
We adopt SSIM to evaluate the reconstruction faithfulness of the generated results. As analyzed in \cite{ZhouWang2004ImageQA}, the structural similarity of two images is measured by three different factors, brightness, contrast, and structure, where the mean is used as the estimate of brightness, the standard deviation as the estimate of contrast, and the covariance as the measurement of structural similarity.

\subsection{Comparison with State-of-the-Art Methods}

\paragraph{Methods.}
We compare our CMVDM with four state-of-the-art (SOTA) methods: MinD-Vis, IC-GANs \cite{ozcelik2022reconstruction}, Gaziv \cite{gaziv2022self}, and Beliy \cite{beliy2019voxels}. We use their official pretrained models for all the comparisons, which are trained on the GOD dataset. For the BOLD5000 dataset, we only compare with the official pretrained MinD-Vis model, because other works \cite{beliy2019voxels, gaziv2022self, ozcelik2022reconstruction} did not conduct experiments and release their models on BOLD5000.

\paragraph{Results on the GOD Dataset.}
We conduct a quantitative comparison between CMVDM and the four SOTA models using the testing dataset of GOD. Table \ref{Comparison Table} summarizes the results, revealing that CMVDM overall outperforms the other methods significantly.
Compared to MinD-Vis and IC-GAN, both of which yield good results, CMVDM outperforms them significantly in terms of SSIM. This indicates that the images generated by CMVDM exhibit a higher degree of resemblance to the visual stimuli in terms of object silhouette and image structure.
Additionally, Fig. \ref{comparison_GOD} demonstrates that CMVDM generates visually impressive images with semantic and structural information closest to the visual stimuli. Gaziv achieves remarkable results in terms of SSIM, but their accuracy reported in Table \ref{Comparison Table} and visual results presented in Fig. \ref{comparison_GOD} demonstrate that their method is not capable of generating high-fidelity images.

\paragraph{Results on the BOLD5000 Dataset.}
We conduct a comparative analysis between our CMVDM and the most recent method MinD-Vis using the testing dataset of BOLD5000. As depicted in Table \ref{Comparison Table}, it is evident that CMVDM consistently outperforms MinD-Vis across all evaluation metrics. Additionally, Fig. \ref{comparison_BOLD5000} provides visualizations of some results from both methods, clearly demonstrating that CMVDM generates more realistic outcomes that are more similar to the GT visual stimuli. Notably, the BOLD5000 dataset, being more complex than the GOD dataset, further validates the effectiveness of our proposed method.

\begin{table}[t]
% \vspace{3 mm}
\centering
\caption{Ablation study of CMVDM's components.}
\label{Ablation Table}
\setlength{\tabcolsep}{4pt}
\begin{tabular}{lccc}
\hline
Method   & Acc (\%)  & PCC & SSIM \\ \hline
MinD-Vis                           & 26.644  & 0.53159  & 0.54489 \\
MinD-Vis+$\mathcal{L}_{align}$     & 27.362  & 0.56686  & 0.52628     \\
MinD-Vis+Control Model             & 28.438  & 0.75730  & \textbf{0.63404}  \\  \hline
% CMVDM w/o cosine loss              & 26.14  & 0.66  & 0.56  \\
CMVDM                              & \textbf{30.112}  & \textbf{0.76751}  & 0.63167  \\ \hline
\end{tabular}
% \vspace{-2 mm}
\end{table}

\begin{figure}[t]
% \vspace{1 mm}
	\centering
	\hspace{-1.5mm}
	\includegraphics[width=0.48\textwidth]{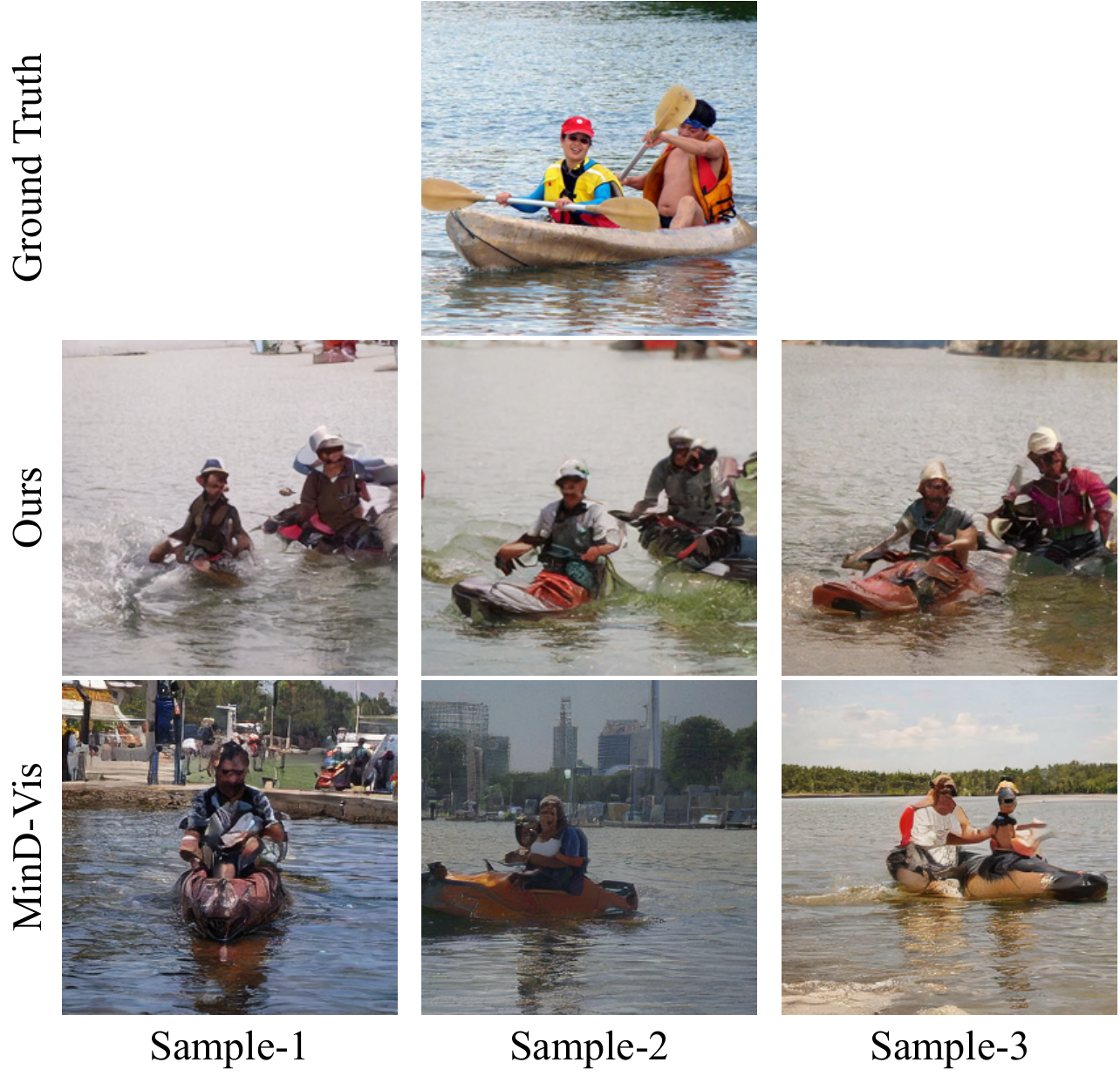} \\
	\caption{Consistency analysis of the generated results.} 
	\label{consistency analysis}
	% \vspace{-5mm}
\end{figure}

\subsection{Ablation Study}

We further conduct experiments on the GOD dataset to analyze the effectiveness of each module of CMVDM. Specifically, we employ MinD-Vis as the baseline and design two comparison models: (1) adding the semantic align loss $\mathcal{L}_{align}$ to MinD-Vis, (2) adding the control model to MinD-Vis. The results, presented in Table \ref{Ablation Table}, demonstrate the efficacy of both $\mathcal{L}_{align}$ and the control model within our CMVDM. MinD-Vis with $\mathcal{L}_{align}$ yields improved results in terms of ACC and PCC, which illustrate that $\mathcal{L}_{align}$ can improve the capability of CMVDM to obtain semantic information. Furthermore, MinD-Vis+Control Model outperforms MinD-Vis+$\mathcal{L}_{align}$ in each metric, particularly in SSIM, indicating that the silhouette contains valuable semantic information that is used in the control model.

\subsection{Consistency Analysis}

To further verify the generative stability of CMVDM, we conduct an analysis to compare the consistency of two diffusion-based methods. As shown in Fig. \ref{consistency analysis}, we sample three images reconstructed by CMVDM and MinD-Vis from the same fMRI signal. The images generated by CMVDM demonstrate a high degree of consistency to GT images both semantically and structurally. However, the results generated by MinD-Vis are capable of reproducing GT images semantically but are not consistent in structure.

\begin{table}[t]
% \vspace{2 mm}
\centering
\caption{Quantitative analysis of the residual block in CMVDM.}
\label{data analysis Table}
\begin{tabular}{ccccc}
\hline
Dataset                   & Method & Acc(\%) & PCC     & SSIM    \\ \hline
\multirow{2}{*}{BOLD5000} & w/o $\mathcal{F}_{res}$ & 25.393 & 0.54184 & 0.52951 \\
                          & w $\mathcal{F}_{res}$   & \textbf{27.791} & \textbf{0.55691} & \textbf{0.53459} \\ \hline
\multirow{2}{*}{GOD}      & w/o $\mathcal{F}_{res}$ & 29.436 & 0.75837 & \textbf{0.63894} \\
                          & w $\mathcal{F}_{res}$   & \textbf{30.112} & \textbf{0.76751} & 0.63167 \\ \hline
\end{tabular}
% \vspace{-3 mm}
\end{table}

\subsection{Further Analysis}
The impact of using the residual module $\mathcal{F}_{res}$ in our CMVDM is significant on the BOLD5000 dataset, as demonstrated in Table \ref{data analysis Table}. However, the effect of $\mathcal{F}_{res}$ on the GOD dataset is not as pronounced. We believe that there are two reasons for this discrepancy.
Firstly, the voxels of a single fMRI signal provided by the BOLD5000 dataset are much less than that provided by the GOD dataset,  making it more challenging to extract valid semantic and silhouette information from BOLD5000. Therefore, $\mathcal{F}_{res}$ is necessary to compensate for the information gap.
Secondly, compared to GOD, BOLD5000 has more diverse images, including scenes that are not present in GOD. The semantic judgment and position alignment of the images in BOLD5000 are more complex than those in GOD. Therefore, we utilize $\mathcal{F}_{res}$ to provide more information and improve the reconstruction performance.
\textit{We provide further investigation on the impact of fMRI signals from different visual cortical regions in the supplementary material.}

\section{Conclusion}
\label{conclusion}

In this paper, we propose a Controllable Mind Visual Diffusion Model (CMVDM) for decoding fMRI signals. Firstly, we simultaneously train a semantic encoder and perform finetuning on a pretrained latent diffusion model to generate semantically consistent images from fMRI signals. Secondly, we incorporate a silhouette extractor to derive reliable position information from fMRI signals. Furthermore, we design a control model to ensure CMVDM generates semantically-consistent and spatially-aligned images with the original visual stimuli. Extensive experiments demonstrate that our approach achieves state-of-the-art performance in generating high-quality images from fMRI signals.

\section{Acknowledgement}
\label{acknowledgement}
This research was supported by Zhejiang Provincial Natural Science Foundation of China under Grant No. D24F020011, Beijing Natural Science Foundation L223024, National Natural Science Foundation of China under Grant 62076016. The work was also supported by 
the National Key Research and Development Program of China (Grant No. 2023YFC3300029) and “One Thousand Plan” projects in Jiangxi Province Jxsg2023102268 and ATR key laboratory grant 220402.

\bibliography{aaai24}

%%%%%%%%%%%%%%%%%%%%%%%%%%%%%%%%%%%%%%%%%%%%%%%%%%%%%%%%%%%%

\newpage

\section*{\LARGE\textbf Appendix}

In this supplementary material, we first give more visualization results, then detail the datasets and the implementation, and finally state the limitations and social impact.

\section{More Visualization Results}

\begin{figure*}
	\centering
	\hspace{-3.5mm}
	\includegraphics[width=0.9\textwidth]{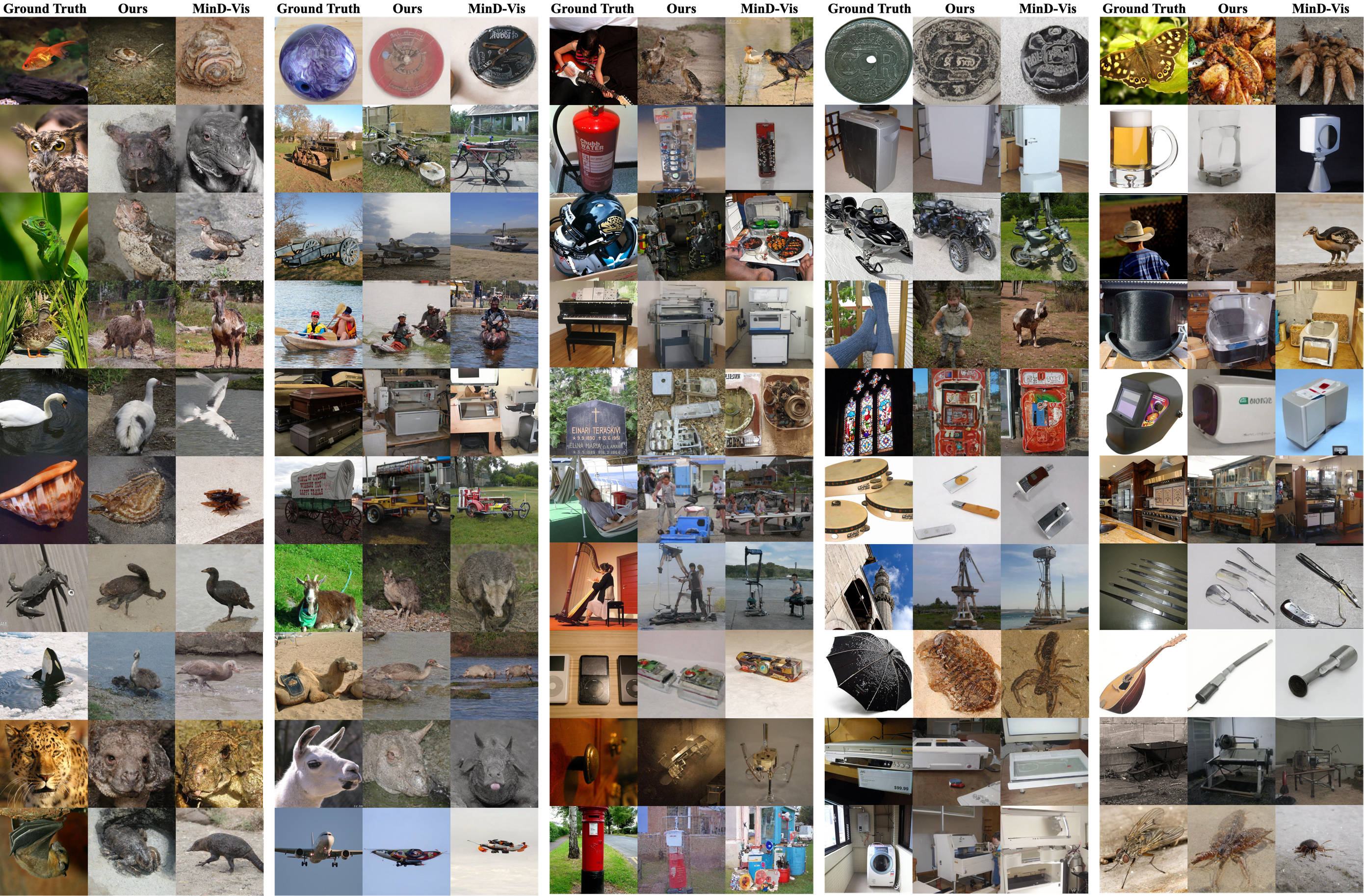} \\
	\caption{Comparison on the GOD Dataset.} 
	\label{comparison_GOD}
\end{figure*}

\begin{figure*}
	\centering
	\includegraphics[width=0.85\textwidth]{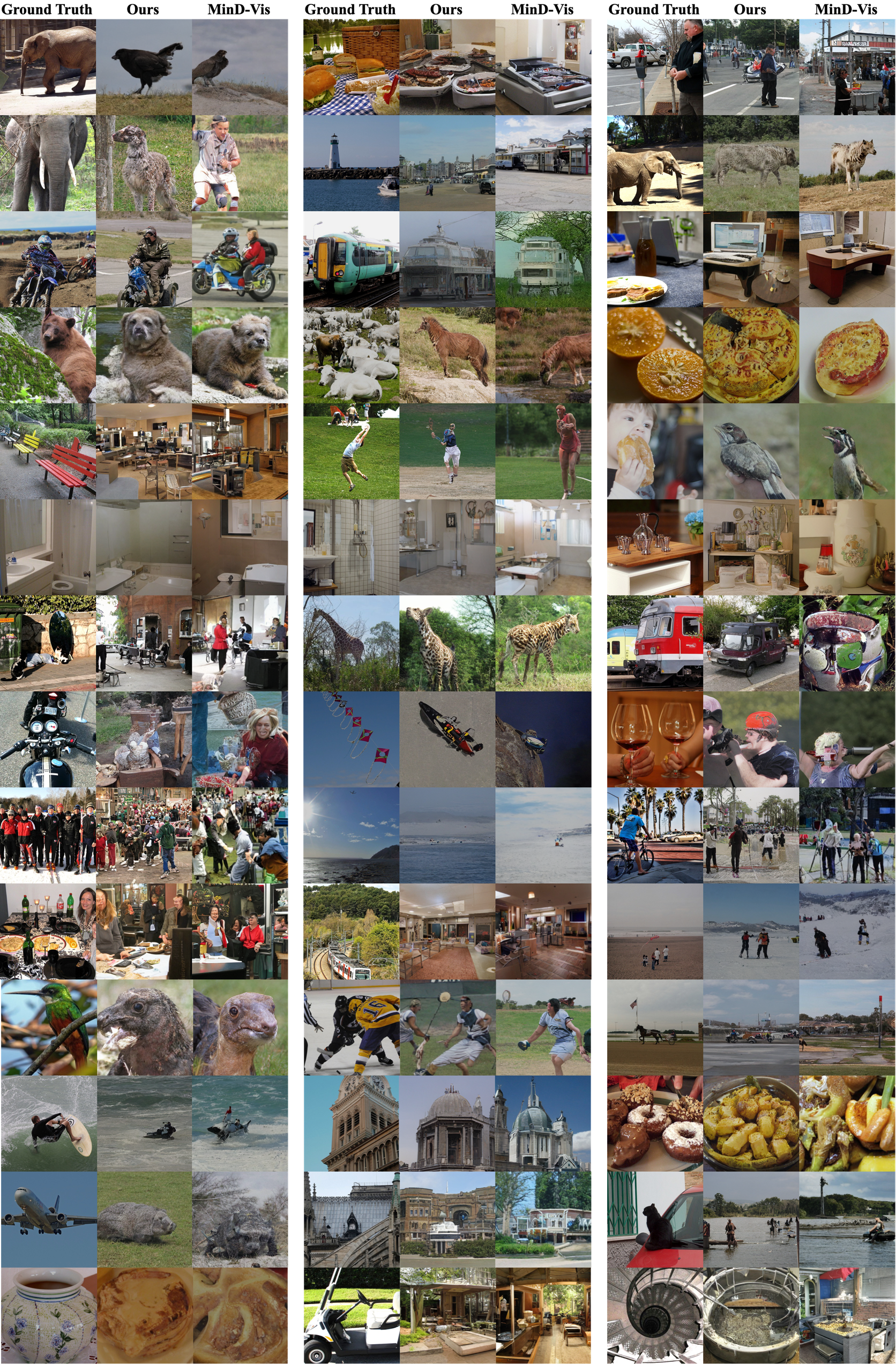} \\
	\caption{Comparison on the BOLD5000 Dataset (Part 1).} 
	\label{comparison_bold1}
\end{figure*}

\begin{figure*}
	\centering
	\includegraphics[width=0.85\textwidth]{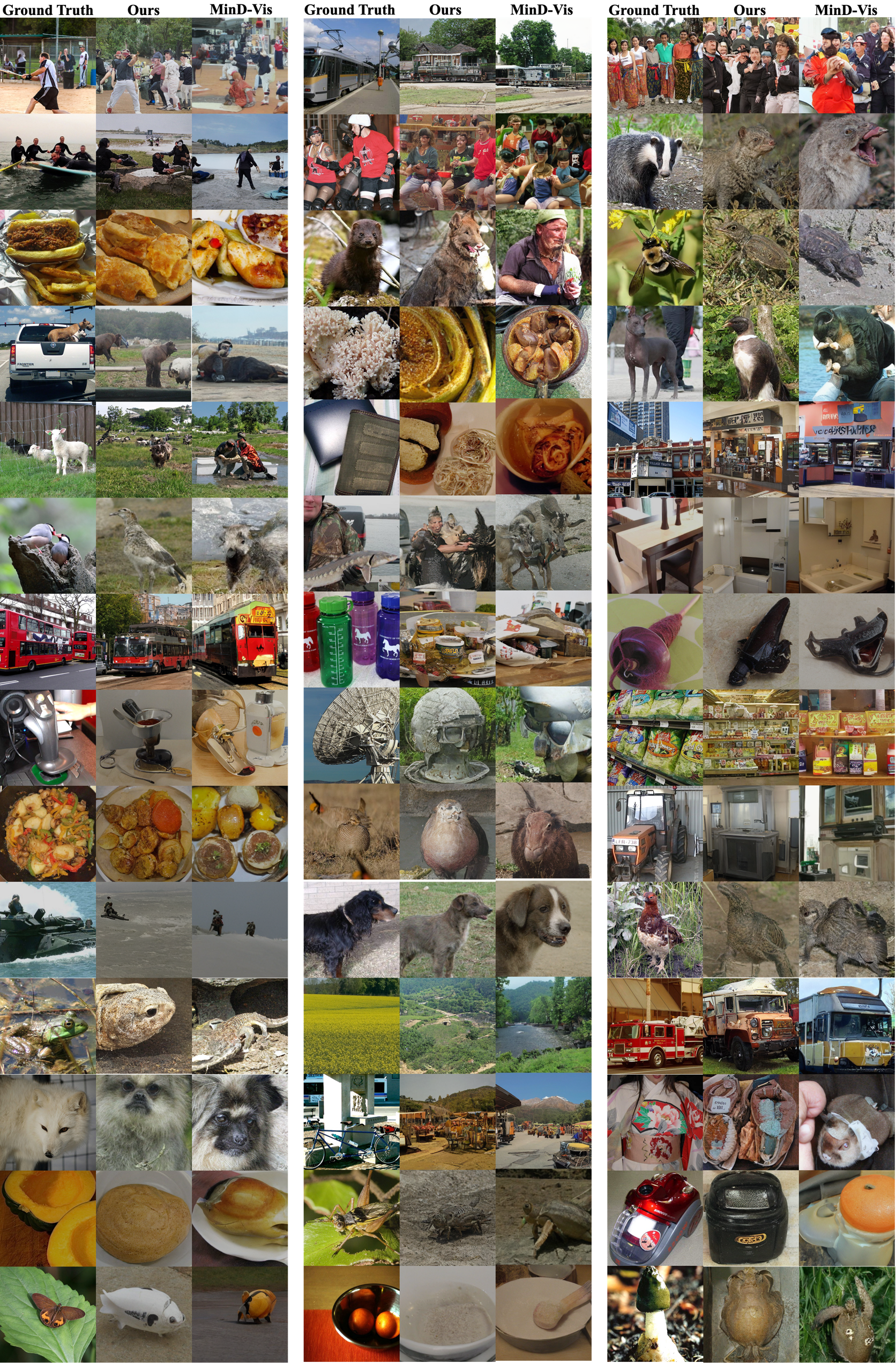} \\
	\caption{Comparison on the BOLD5000 Dataset (Part 2).} 
	\label{comparison_bold2}
\end{figure*}

\begin{figure*}
	\centering
	\includegraphics[width=0.85\textwidth]{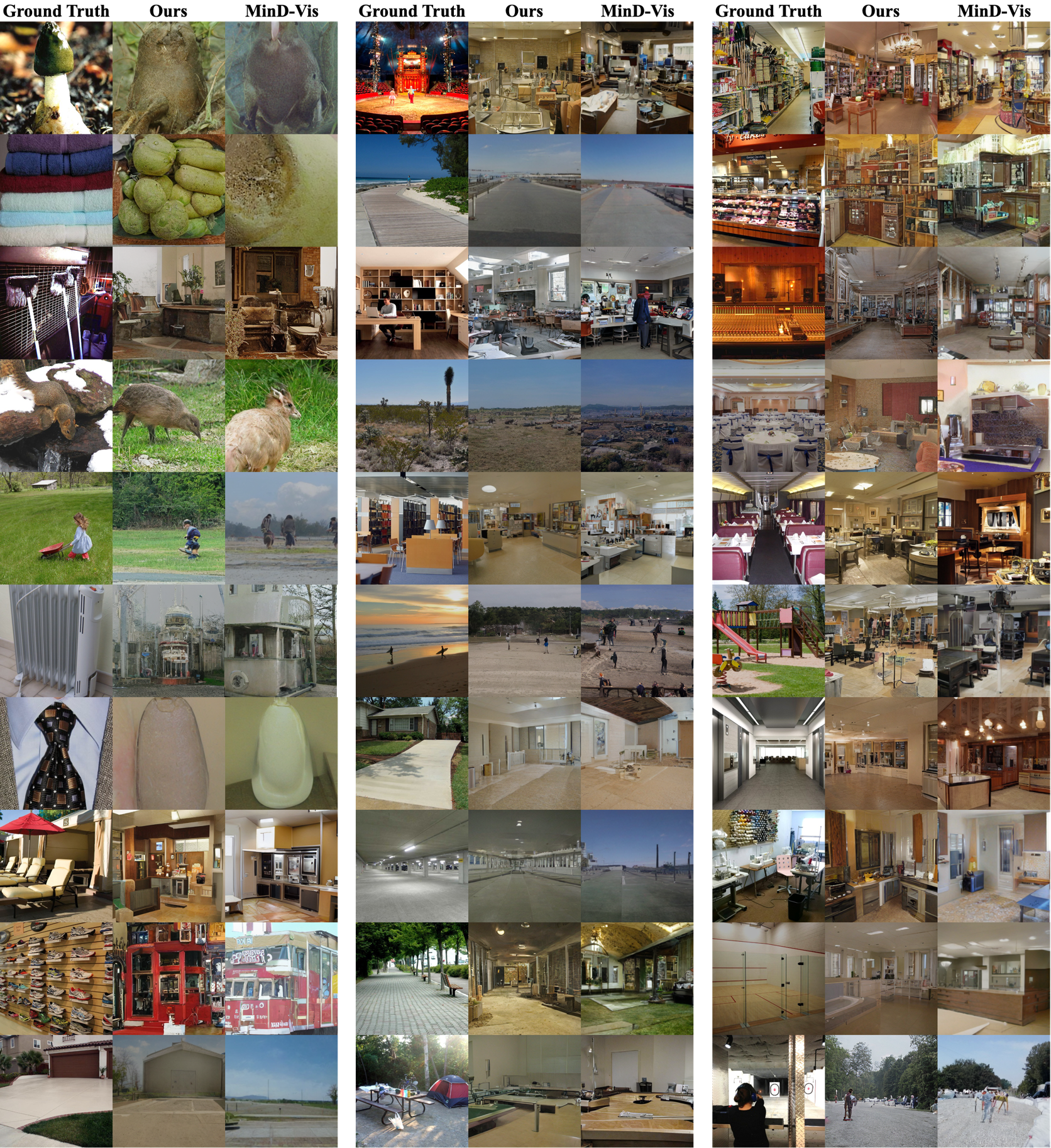} \\
	\caption{Comparison on the BOLD5000 Dataset (Part 3).} 
	\label{comparison_bold3}
\end{figure*}

In this section, we present comprehensive visualizations of all the samples from the test sets of the BOLD5000 and GOD datasets. Each group consists of three columns, with each representing the original visual stimuli (Ground Truth) from the test dataset, the output generated by CMVDM, and the results generated by MinD-Vis \cite{chen2022seeing}. The visualization of the 50 images from the GOD test set is shown in Fig. \ref{comparison_GOD}, and the visualization of the 113 images from the BOLD5000 test set is illustrated in Figs. \ref{comparison_bold1}, \ref{comparison_bold2}, \ref{comparison_bold3}

\section{Datasets}

\paragraph{Generic Object Decoding Dataset.}

The Generic Object Decoding (GOD) \cite{horikawa2017generic} dataset is a collection of $1250$ paired \{fMRI, image\} data. The images from \cite{deng2009imagenet} were presented to five healthy subjects, while an fMRI scanner recorded their brain activity. The dataset was collected in two sessions. In the training session, $1,200$ images from $150$ categories were presented to each subject only once. In the test session, $50$ images from $50$ categories were presented $35$ times each. Following previous works \cite{chen2022seeing,ozcelik2022reconstruction,gaziv2022self,beliy2019voxels}, we chose the data from Subject 3 in the GOD dataset to make a fair comparison. We computed the average of the fMRI signals across all trials to arrive at the final results for evaluation. Notably, there was no overlap between the categories of images shown to the subjects in the training session and the test session. After the scanning sessions, the acquired fMRI data underwent motion correction and then was co-registered to the whole-head high-resolution anatomical images. By conducting standard retinotopy and localizer experiments, the authors determined $7$ brain regions (V1, V2, V3, V4, LOC, FFA, PPA) and combined their voxels to define the entire visual cortex (VC). We selected the voxels from VC ($4643$ voxels in total) as the fMRI data to conduct our experiments.

\paragraph{Brain, Object, Landscape Dataset.}

The Brain, Object, Landscape Dataset (BOLD5000) has $5,254$ \{fMRI,image\} pairs from $4,916$ unique images from Scene UNderstanding (SUN) \cite{xiao2010sun}, Common Objects in Context (COCO) \cite{lin2014microsoft} and ImageNet \cite{deng2009imagenet}. In this dataset, $4,803$ training images and 113 test images were presented to $4$ subjects aged between $24$ and $27$. BOLD5000 alleviates the issue of limited data in the field and encompasses a broad variety of image categories, while also enriching the visual stimuli with real-world indoor and outdoor scenes. This enables a detailed investigation into the neural representation of visual inputs across a wide range of visual semantics. 
It is noteworthy that the BOLD5000 dataset contains a smaller number of voxels per subject ($1685$) than that in the GOD dataset ($4643$), due to the difference in the fMRI scanners used to collect the data.

\paragraph{Human Connectome Project fMRI Dataset.}

The Human Connectome Project (HCP) \cite{van2013wu} is a large-scale research project that aims to map the structural and functional connectivity of the human brain using advanced neuroimaging techniques. One of the main datasets produced by HCP is the fMRI dataset, which measures the blood oxygen level-dependent (BOLD) signal changes in response to various tasks and resting states. The fMRI dataset consists of high-resolution ($2$ mm isotropic) data from $1200$ healthy young adults (ages $22-35$). The dataset includes four sessions of resting-state fMRI (rs-fMRI), and seven sessions of task-based fMRI (T-fMRI), covering motor, working memory, language, social, relational, emotion, and gambling domains. The fMRI data have been extensively preprocessed and analyzed using state-of-the-art methods, such as independent component analysis (ICA), seed-based correlation analysis (SCA), and multivariate pattern analysis (MVPA). 
Following MinD-Vis, we pretrain the fMRI encoder $\mathcal{E}_{fmri}$ on this dataset to learn effective fMRI representations.

\paragraph{ImageNet Validation Dataset.}
The ImageNet Validation Dataset is a subset of ImageNet \cite{deng2009imagenet}, which is a large-scale visual database containing millions of images annotated with object labels. The ImageNet Validation Dataset consists of $50,000$ images, each labeled with one of $1000$ different object categories. These categories are organized hierarchically according to the WordNet \cite{miller1998wordnet} hierarchy, and each category may correspond to multiple words or phrases. The ImageNet Validation Dataset is commonly used to evaluate the performance of machine learning models on the image classification task.
In our method, we utilized the ImageNet Validation Dataset as the dataset for self-supervised training of $\mathcal{D}_{slh}$, which aims to reconstruct images that are structurally similar to the original visual stimuli. The dataset covers a wide range of object categories, such as animals, plants, vehicles, furniture, clothing, and more. This diversity makes it suitable for learning generalizable and diverse features.

\section{More Implementation Details}

\subsection{Architecture of the Silhouette Estimation Network}
The silhouette estimation network consists of two components, an encoder $\mathcal{E}_{slh}$ and a decoder $\mathcal{D}_{slh}$. 
The encoder $\mathcal{E}_{slh}$ is used for encoding natural images and projecting the image features to the fMRI signal space. Following \cite{gaziv2022self}, $\mathcal{E}_{slh}$ first extracts image features using a pretrained VGG19 network. The features are then fed into downsampling blocks with batch normalization, $\times 2$ maximum pooling, $3 \times 3$ convolution with $32$ channels, ReLU activation, and batch normalization to obtain the hierarchy of semantic and spatial representations. Finally, the four representations are concatenated and mapped into the fMRI signal space by a full-connected layer. 

The decoder $\mathcal{D}_{slh}$ architecture uses a full-connected layer to transform and reshape the fMRI input into a $64 \times 14 \times 14$ feature map. This feature map is then fed into three blocks, and each consists of $\times 2$ up-sampling, $5 \times 5$ convolution with $64$ channels, ReLU activation, and group normalization. A convolution layer is employed to output an RGB silhouette image.

\subsection{Architecture of the LDM}

The Latent Diffusion Model (LDM) in this work consists of an autoencoder and a denoising network with a UNet structure. The autoencoder's encoder and decoder both have a depth of $3$ blocks. The blocks of the encoder have feature channel sizes of $128$, $256$, and $512$, respectively, while the decoder follows the same structure in reverse, and the latent space of the autoencoder corresponds to a feature resolution of $64  \times 64$. 
In addition, the denoising network of the LDM employs a UNet architecture, featuring an encoder and a decoder, each comprising four blocks in depth.
The encoder blocks have channel sizes of $192$, $384$, $576$, and $960$, respectively. Besides, the LDM takes a $512$-dimensional condition input.

At the same time, the control model primarily consists of a hint model which is employed to process the input condition, an encoder, and several zero-convolution layers. The hint model is composed of an 8-layer convolutional neural network, and the structure of the encoder of the control model is identical to that of the encoder in the LDM's denoising network.

\subsection{Architechture of the Residual Block}

The residual model, denoted as $\mathcal{F}_{res}$, contains a Multilayer Perceptron (MLP) configured with three fully connected (FC) layers and a convolutional neural network (CNN) with four distinct layers. The output provided by this model is added to the output generated by $\mathcal{D}_{slh}$.

\subsection{Evaluation Metric}

The specific steps for N-way Classification Accuracy (Acc) are as follows: firstly, we use a pre-trained classifier to obtain the ground truth class $\hat{y}$. Then, we calculate the classification results for the generated image using the classifier $p={p_0,...,p_{999}}$. Next, we randomly select n-1 classes and combine them with the ground truth class to obtain a new set of probabilities $p^\prime={p_{\hat{y}},p_{y_{1}},...,p_{y_{n-1}}}$. If the max value of $p^\prime_y$ is $p_{\hat{y}}$, we consider it as a correct classification; otherwise, it is an incorrect classification. We repeat this process multiple times on the entire testing dataset to calculate the classification accuracy.

\begin{figure*}
	\centering
	\hspace{-3.5mm}
	\includegraphics[width=0.98\textwidth]{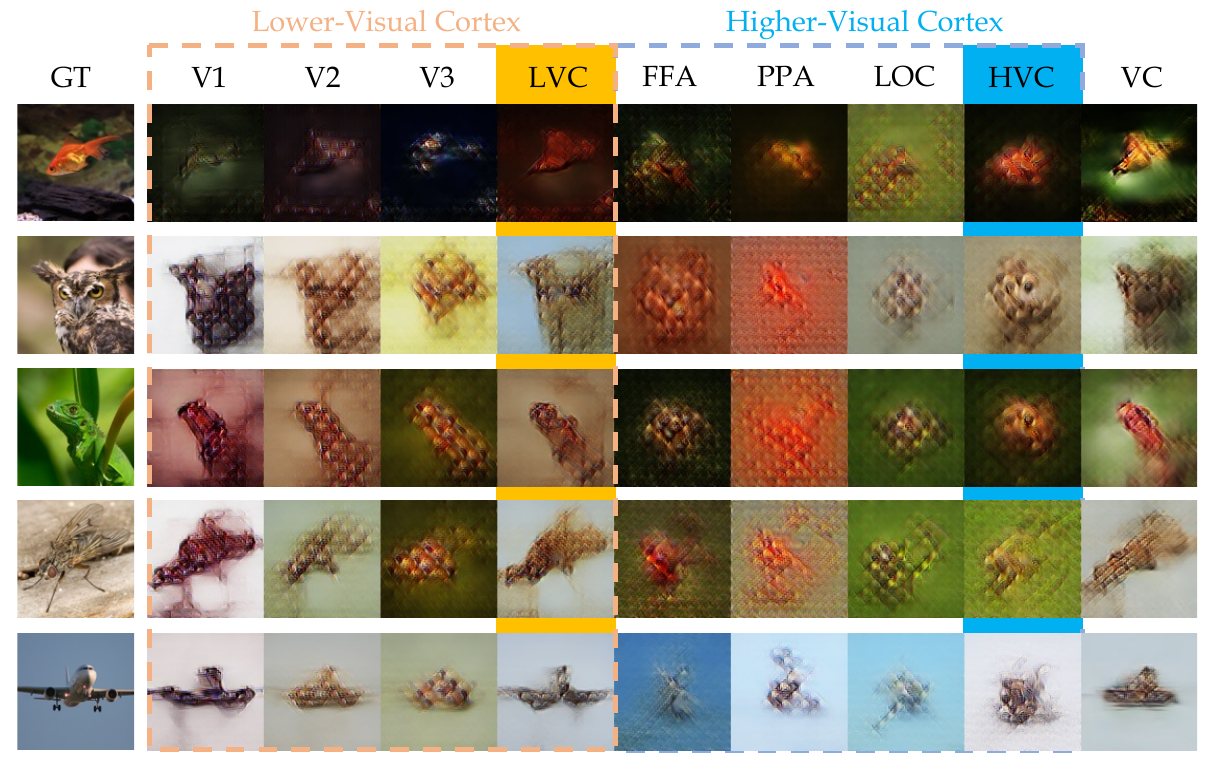} \\
	\caption{Comparison of silhouettes reconstructed from fMRI voxels of different visual regions. The primary components of the lower visual cortex include V1, V2, and V3, with LVC representing all voxels of the lower visual cortex. Similarly, the higher visual cortex primarily comprises FFA, PPA, and LOC, where HVC represents all voxels of the higher visual cortex. VC represents the voxels of the entire visual cortex.} 
	\label{comparison_GOD_slh}
\end{figure*}

\begin{figure*}
	\centering
	\hspace{-3.5mm}
	\includegraphics[width=0.75\textwidth]{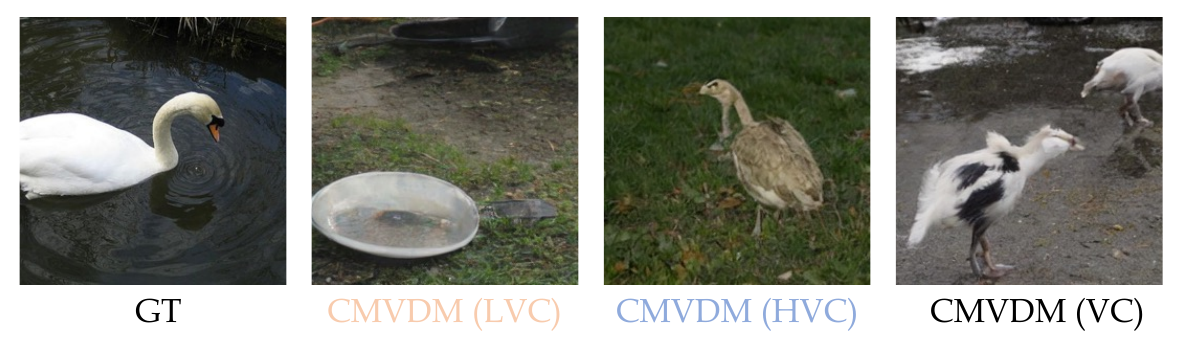} \\
	\caption{Reconstructed images from different regions of the visual cortex by our method.} 
	\label{comparison_LVC}
\end{figure*}

\subsection{Additional Analysis}

\begin{table}[]
\centering
\caption{Comparison of the performance between LVC, HVC, and VC regions using MinD-Vis and CMVDM, against the test dataset of GOD.}
\begin{tabular}{l|ccc}
\hline
                 & Acc (\%) & PCC  & SSIM \\ \hline
MinD-Vis $\mathcal{L}_{align}$ (LVC) & 6.80     & 0.57 & 0.47 \\
MinD-Vis $\mathcal{L}_{align}$ (HVC) & 17.20    & 0.50 & 0.45 \\
MinD-Vis $\mathcal{L}_{align}$ (VC)  & 27.36    & 0.57 & 0.53 \\ \hline
CMVDM (LVC)    & 8.80     & 0.68 & 0.63 \\
CMVDM (HVC)    & 21.62    & 0.55 & 0.53 \\
CMVDM (VC)     & \textbf{30.11}    & \textbf{0.77} & \textbf{0.63} \\ \hline
\end{tabular}
\label{regions}
\end{table}

We have further investigated the impact of fMRI signals from different visual cortical regions on our experimental outcomes. This investigation has two main components: probing the quality of silhouette reconstruction based on voxels from distinct visual cortical areas and analyzing the effects of different hierarchical visual cortices (higher visual cortex (HVC) and lower visual cortex (LVC)) on the semantic and positional aspects of the reconstruction results.

Initially, leveraging the GOD dataset, we explore the influence of fMRI signals from various visual cortical areas (V1, V2, V3, FFA, PPA, LOC, LVC, HVC, and VC). These input signals are individually utilized for pretraining silhouette decoder $\mathcal{D}_{slh}$ and applied to reconstruct images from the test set fMRI signals using $\mathcal{D}_{slh}$. The comparative visualization results are presented in Fig. \ref{comparison_GOD_slh}. Notable observations include:
\begin{itemize}
    \item V1, V2, and V3, as components of the LVC, yield reconstructed images with spatial structures that closely resemble ground truth (GT) images. This strongly suggests the role of the lower visual cortex in processing spatial information in visual signals.
    
    \item The visualizations of FFA, PPA, and LOC (from HVC) lack interpretable spatial structures. However, these regions are hypothesized to have meaningful semantic comprehension of visual signals, which is further validated by subsequent experiments.
    
\end{itemize}

In our paper, we use the VC signal that encompasses signals from all visual cortical areas. The results of its reconstruction are also shown in Fig. \ref{comparison_GOD_slh}. Compared to the images reconstructed solely using signals from individual regions, the VC images capture both the structural and RGB information that closely aligns with the GT image distribution. This richness of information facilitates more accurate control over high-fidelity image generation.

We then train our CMVDM model using fMRI signals from the LVC, HVC, and VC regions. For reference, we also train MinD-Vis with the $\mathcal{L}_{align}$ under the same setting. We use the same evaluation metrics as in the paper, and the results are presented in Table \ref{regions}. Notable observations include:

\begin{itemize}
    \item Both methods exhibit lower Acc on LVC, which suggests that the semantic information on LVC is unclear. However, they both demonstrate high structural similarity (PCC and SSIM), affirming the functional interpretation of fMRI signals within the LVC region.

    \item Both methods perform better on HVC with higher Acc, signifying increased semantic similarity. However, PCC and SSIM, measuring structural similarity, are lower, suggesting that the HVC region contributes vital semantic information to the visual comprehension process.

    \item VC, encompassing voxels from both LVC and HVC, achieves the highest values in both semantic and structural similarity metrics. This implies the effective and substantial fusion of information from high and low visual cortical regions in our CMVDM approach.

\end{itemize}

Furthermore, we visually compare the reconstruction results of CMVDM trained on LVC, HVC, and VC regions against the test set in Fig. \ref{comparison_LVC}. This visualization corroborates the results presented in Table \ref{regions} to some extent.

\section{Discussion and Limitations.}

Our visualization results demonstrate that CMVDM generates better images on the GOD dataset than those on the BOLD5000 dataset in terms of image structure and object silhouette. This difference can be attributed to the silhouette information $c_{slh}$ extracted by $\mathcal{D}_{slh}$ being less satisfactory on the BOLD5000 data. Two possible reasons for this are as follows: Firstly, the BOLD5000 dataset is more complex due to the greater diversity of indoor/outdoor scenes, as well as interactions between objects, including both single and multiple objects. On the other hand, the GOD dataset only focuses on single objects. 
Secondly, due to the difference in the fMRI scanners and experimentation settings, the acquisition and preprocessing procedures for fMRI signals may vary. Specifically, a single fMRI signal in the BOLD5000 dataset contains fewer voxels (1685) compared to the GOD dataset (4643), which increases the difficulty of extracting meaningful semantic and positional information using the fMRI encoder $\mathcal{E}_{fmri}$. These factors may pose challenges to generating satisfactory $c_{slh}$ on the BOLD5000 dataset. 

While CMVDM outperforms prior approaches in generating more plausible results, it exhibits a  discrepancy between the two datasets. This may be due to the small sizes of the two datasets we used, which prevent our CMVDM from being verified sufficiently. Therefore, a potential limitation of this study is the lack of validation on a larger paired fMRI-image dataset. Additionally, as mentioned above, the fMRI signals obtained under different experimental conditions vary a lot, and the cross-domain generation ability and robustness of the model still need to be further explored.  We plan to address these limitations and further improve our approach in future studies.

\section{Social Impact}

This work does not have a direct negative social impact. However, we should pay attention to the ethical and privacy issues in the process of collecting or using our model to visualize fMRI signals and prevent them from being abused for malicious purposes.

\end{document}